\documentclass[11pt,a4paper]{article}
\usepackage[utf8x]{inputenc} 
\usepackage[T1]{fontenc}    
\usepackage{lmodern}
\usepackage{url}            
\usepackage{booktabs}       
\usepackage{amsfonts, amssymb, amsmath, amsthm}       
\usepackage{nicefrac}       
\usepackage{microtype}      
\usepackage{algorithm}
\usepackage{algorithmic}
\usepackage{color}
\usepackage{graphicx}
\usepackage{geometry}
\usepackage{bm}
\usepackage{enumitem}
\usepackage{fancyvrb}
\usepackage{caption}
\usepackage{wrapfig}
\usepackage{cleveref}
\usepackage{multirow}
\usepackage{bbm}
\usepackage{subcaption} 

\newcommand{\deq}{\mathrel{\mathop:}=}

\newcommand{\E}{\mathbb{E}}
\newcommand\trans{^\top}
\renewcommand{\epsilon}{\varepsilon}
\newcommand{\norm}[1]{\left\lVert#1\right\rVert}

\newcommand{\sbar}[1]{\tilde{s}_{#1}}
\newcommand{\thetabar}[1]{\tilde{\theta}_{#1}}
\newcommand{\gbar}[1]{\tilde{g}_{#1}}
\newcommand{\real}{\mathbb{R}}
\newcommand{\SGDOpt}{\mathrm{SGDOpt}}
\newcommand{\Fout}{F_\mathrm{out}}
\newcommand{\Fstate}{F_\mathrm{state}}
\newcommand{\comment}[1]{\textbf{/* #1 */}}
\newcommand{\target}[1]{#1^\ast} 

\newtheorem{proposition}{Proposition}

\title{Unbiased Online Recurrent Optimization}

\date{}
\author{Corentin Tallec, Yann Ollivier}

\begin{document}

\maketitle

\begin{abstract}
The novel \emph{Unbiased Online Recurrent Optimization}
(UORO) algorithm allows for online learning of general recurrent
computational graphs such as recurrent network models. It works in a
streaming fashion and avoids backtracking through past activations and
inputs. UORO is computationally as costly as \emph{Truncated
Backpropagation Through Time} (truncated BPTT), a widespread algorithm for online
learning of recurrent networks \cite{jaeger2002tutorial}. 
UORO is a modification of \emph{NoBackTrack} 
\cite{DBLP:journals/corr/OllivierC15} that bypasses the need for model
sparsity and makes implementation easy in current deep learning
frameworks, even for complex models.  

Like NoBackTrack,
UORO provides unbiased gradient estimates; unbiasedness is the
core hypothesis in 
stochastic
gradient descent theory, without which
convergence to a local optimum is not guaranteed. On the contrary,
truncated BPTT does not provide this property, leading
to possible divergence.

On synthetic tasks where truncated BPTT is shown to diverge, UORO converges. For
    instance, when a parameter has a positive short-term but negative long-term
    influence, truncated BPTT diverges unless the truncation span is very
    significantly longer than the intrinsic temporal range of the
    interactions, while UORO performs well thanks to the unbiasedness of its
    gradients.
\end{abstract}

Current recurrent network learning algorithms are ill-suited to online learning
via a single pass through long sequences of temporal data.
\emph{Backpropagation Through Time} (BPTT \cite{jaeger2002tutorial}), the
current
standard for training recurrent architectures, is well suited to many short
training sequences. Treating long sequences with BPTT requires either
storing all past inputs in memory and 
waiting for a long time between each learning step,
or arbitrarily splitting the
input sequence into smaller sequences, and applying BPTT to each of those
short sequences, at the cost of losing long term dependencies.

This paper introduces \emph{Unbiased Online Recurrent Optimization}
(UORO), an \emph{online} and \emph{memoryless} learning algorithm for
recurrent architectures: UORO processes and learns from data samples
sequentially, one sample at a time. Contrary to BPTT, UORO does not
maintain a history of previous inputs and activations. Moreover, UORO is
\emph{scalable}: processing data samples with UORO comes at a similar
computational and memory cost as just running the recurrent model on
those data.

Like most neural network training algorithms, UORO relies on stochastic
gradient optimization. The theory of stochastic gradient crucially
relies on the unbiasedness of gradient estimates to provide convergence to
a local optimum. To this end, in the footsteps of \emph{NoBackTrack}
(NBT)
\cite{DBLP:journals/corr/OllivierC15}, UORO provides
provably \emph{unbiased} gradient estimates, in a scalable, streaming fashion.

Unlike NBT, though, UORO can be easily implemented in a black-box fashion on
top of an existing recurrent model in current machine learning software,
without delving into the mathematical structure and code of the model.

The framework for recurrent optimization and UORO is introduced in
Section~\ref{sec:back-forw}.
The final algorithm is reasonably simple (Alg.~\ref{alg:uoro}), but its
derivation (Section~\ref{sec:uoro}) is more complex.
In
Section~\ref{sec:exp}, UORO is shown to provide convergence on a set of
synthetic experiments where truncated BPTT fails to display reliable
convergence.  An implementation of UORO is provided \href{https://github.com/ctallec/uoro}{here}.

\section{Related work}
A widespread approach to online learning of recurrent neural networks is
\emph{Truncated Backpropagation Through Time} (truncated BPTT)
\cite{jaeger2002tutorial}, which mimics
Backpropagation Through Time, but zeroes gradient flows after a fixed number of
timesteps. This truncation makes gradient estimates biased; consequently,
truncated BPTT does not provide any convergence guarantee. Learning is  biased towards
short-time dependencies. 
\footnote{ Arguably, truncated BPTT might still learn some dependencies beyond its
truncation range, by a mechanism similar to Echo State Networks
\cite{jaeger2002tutorial}. However, truncated BPTT's gradient estimate has a marked
bias towards short-term rather than long-term dependencies, as shown in the
first experiment of Section~\ref{sec:exp}.  }. Storage of some past
inputs and states is required. 

Online, exact gradient computation methods have long been known (\emph{Real Time
Recurrent Learning} (RTRL)
\cite{Williams:1989:LAC:1351124.1351135,pearlmutter1995gradient}), but their computational cost
discards them for reasonably-sized networks.

\emph{NoBackTrack} (NBT) \cite{DBLP:journals/corr/OllivierC15} also provides
unbiased gradient estimates for recurrent neural networks. However, contrary
to UORO, NBT cannot be applied in a blackbox fashion, making it extremely
tedious to implement for complex architectures.

Other previous attempts to introduce generic online learning algorithms with
a
reasonable computational cost all result in biased gradient estimates.
\emph{Echo State Networks} (ESNs) \cite{jaeger2002tutorial,jaegerESN} simply set to $0$ the gradients
of recurrent parameters. Others, e.g., \cite{Maass2002,steil04backpropagation}, introduce approaches
resembling ESNs, but keep a partial estimate of the recurrent gradients.
The original \emph{Long Short
Term Memory} algorithm \cite{Hochreiter:1997:LSM:1246443.1246450} (LSTM now refers to a particular
architecture) cuts gradient flows going out of gating units
to make gradient computation tractable. \emph{Decoupled Neural Interfaces}
\cite{DBLP:journals/corr/JaderbergCOVGK16} bootstrap
truncated gradient estimates using synthetic gradients generated by
feedforward neural networks.
The algorithm in \cite{Movellan} provides zeroth-order estimates of recurrent
gradients via diffusion networks; it could arguably be turned online by running
randomized alternative trajectories. Generally
these approaches lack a strong theoretical backing, except arguably ESNs.

\section{Background}
\label{sec:back-forw}

UORO is a learning algorithm for recurrent computational graphs.
Formally, the aim is to optimize $\theta$, a parameter
controlling the evolution of a dynamical system 
\begin{align} 
    s_{t+1} &=
    \Fstate(x_{t+1}, s_t, \theta) \label{eq:state-eq}\\ 
    o_{t+1} &= \Fout(x_{t+1}, s_t, \theta)
\end{align} 
in order to minimize a total loss $\mathcal{L} \deq \sum\limits_{0\leq
t\leq T} \ell_t(o_t,
\target{o}_t)$, where $\target{o}_t$ is a target output at time $t$. 
For instance, a standard recurrent neural network, with hidden state
$s_t$ (preactivation values) and output $o_t$ at time $t$,
is described
with the update equations $\Fstate(x_{t+1}, s_t,
\theta)\deq W_x\,x_{t+1} + W_s\,\text{tanh}(s_t) + b$ and $\Fout(x_{t+1}, s_t,
\theta) \deq W_o\,\text{tanh}(\Fstate(x_{t+1}, s_t, \theta)) + b_o$; here the parameter is 
$\theta=(W_x, W_s,
b, W_o, b_o)$, and a typical loss might be $\ell_s(o_s,
\target{o}_s)\deq (o_s-\target{o}_s)^2$.

Optimization by gradient descent is standard for neural networks.
In the spirit of stochastic gradient descent, we can optimize the total loss
$\mathcal{L} = \sum\limits_{0\leq t\leq T} \ell_t(o_t,
\target{o}_t)$
one term at a time and 
update the parameter online at each time step via
\begin{equation} 
    \theta \leftarrow \theta - \eta_t \frac{\partial \ell_t}{\partial \theta}\trans
    \label{eq:sgd}
\end{equation}
where $\eta_t$ is a scalar learning rate at time $t$. (Other
gradient-based optimizers can also be used, once $\frac{\partial
\ell_t}{\partial \theta}$ is known.) The focus is then to
compute, or approximate, $\frac{\partial \ell_t}{\partial \theta}$.

BPTT computes $\frac{\partial \ell_t}{\partial \theta}$ by unfolding
the network through time, and backpropagating through the unfolded
network, each timestep corresponding to a layer. BPTT thus requires
maintaining the full unfolded network, or, equivalently, the history of
past inputs and activations.  \footnote{Storage of past activations can
be reduced, e.g. \cite{membp}. However, storage of all past inputs is
necessary.} \emph{Truncated BPTT}
only unfolds the network for a fixed number of
timesteps, reducing computational cost in online settings
\cite{jaeger2002tutorial}.
This comes at the cost of biased gradients, and can prevent
convergence of the gradient descent even for large truncations, as
clearly exemplified in Fig.~\ref{fig:bpttf}.

\section{Unbiased Online Recurrent Optimization}
\label{sec:uoro}
Unbiased Online Recurrent Optimization is built on top of a forward computation of the
gradients, rather than backpropagation. Forward gradient computation
for neural networks (RTRL) is described in
\cite{Williams:1989:LAC:1351124.1351135} and we review it in
Section~\ref{sec:rtrl}. The derivation of UORO
follows in Section~\ref{sec:rk1}. Implementation
details are given in Section~\ref{sec:impl}. UORO's derivation
is strongly connected to \cite{DBLP:journals/corr/OllivierC15} but
differs in one critical aspect:
the sparsity hypothesis made in the latter is relieved, resulting in
reduced implementation complexity without any model restriction.

\subsection{Forward computation of the gradient}
\label{sec:rtrl}
Forward computation of the gradient for a recurrent model (RTRL)
is directly obtained by applying
the chain rule to both the loss function and the state equation
\eqref{eq:state-eq}, as follows.

Direct differentiation and application of the chain rule to $\ell_{t+1}$ yields
\begin{equation}
    \frac{\partial \ell_{t+1}}{\partial \theta}
    =\frac{\partial \ell_{t+1}}{\partial o}(o_{t+1}, \target{o}_{t+1})\cdot\left(\frac{\partial \Fout}{\partial s}(x_{t+1}, s_t, \theta)\,\frac{\partial s_t}{\partial \theta} + 
    \frac{\partial \Fout}{\partial \theta}(x_{t+1}, s_t, \theta)\right).
    \label{eq:lossgrad}
\end{equation}

Here, the term $\partial s_t/\partial \theta$ represents the effect on
the state at time $t$ of a
change of parameter during the whole past trajectory.
This term can be computed inductively from time $t$ to $t+1$. 
Intuitively, looking at the update equation \eqref{eq:state-eq}, there are two contributions to $\partial s_{t+1}/\partial \theta$:
\begin{itemize}
    \item The direct effect of a change of $\theta$ on the computation of
    $s_{t+1}$, given $s_t$.
    \item The past effect of $\theta$ on $s_t$ via the whole past
    trajectory.
\end{itemize}
With this in mind, differentiating \eqref{eq:state-eq} with respect to
$\theta$ yields
\begin{equation}
    \frac{\partial s_{t+1}}{\partial \theta} = \frac{\partial \Fstate}{\partial \theta}(x_{t+1}, s_t, \theta) + 
\frac{\partial \Fstate}{\partial s}(x_{t+1}, s_t, \theta)\,\frac{\partial s_t}{\partial \theta}.
    \label{eq:rtrl}
\end{equation}

This gives a way to compute the derivative of the instantaneous loss
without storing past history: at each time step, update $\partial
s_{t}/\partial \theta$ from $\partial s_{t-1}/\partial \theta$, then use this
quantity to directly compute $\partial \ell_{t+1}/\partial \theta$. This is
how RTRL \cite{Williams:1989:LAC:1351124.1351135} proceeds.

A huge disadvantage of RTRL is that $\partial s_t/\partial \theta$ is
of size $\text{dim}(\text{state}) \times \text{dim}(\text{params})$.  For
instance, with a fully connected standard recurrent network with $n$
units, $\partial s_t/\partial \theta$ scales as $n^3$. This makes RTRL
impractical for reasonably sized networks.

UORO modifies RTRL by only maintaining a scalable,
rank-one, provably unbiased approximation of $\partial s_t/\partial \theta$, to reduce the memory
and computational cost. This approximation takes the form
$\sbar{t}\otimes\thetabar{t}$, where $\sbar{t}$ is a column vector of the same
dimension as $s_t$, $\thetabar{t}$ is a \emph{row} vector of the same dimension as
$\theta\trans$, and $\otimes$ denotes the outer product. The resulting quantity is
thus a matrix of the same size as $\partial s_t/\partial \theta$. The
memory cost of storing $\sbar{t}$ and $\thetabar{t}$ scales as
$\text{dim}(\text{state})+\text{dim}(\text{params})$.  Thus UORO is as
memory costly as simply running the network itself (which indeed requires
to store the current
state and parameters). The following section details how $\sbar{t}$ and $\thetabar{t}$ are
built to provide unbiasedness.

\subsection{Rank-one trick: from RTRL to UORO}
\label{sec:rk1}

Given an unbiased estimation of $\partial s_t/\partial \theta$, namely, a
stochastic matrix $\tilde{G}_t$ such that $\mathbb{E}\,\tilde{G}_t =
\partial s_t/\partial \theta$, unbiased estimates of $\partial
\ell_{t+1}/\partial \theta$ and $\partial s_{t+1}/\partial \theta$ can be
derived by plugging $\tilde{G}_t$ in \eqref{eq:lossgrad} and
\eqref{eq:rtrl}. Unbiasedness is preserved thanks to linearity of the mean, because both
\eqref{eq:lossgrad} and \eqref{eq:rtrl} are affine in $\partial
s_t/\partial \theta$.

Thus, assuming the existence of a rank-one unbiased approximation $\tilde{G}_t
= \sbar{t} \otimes \thetabar{t}$ at time $t$, we
can plug it in \eqref{eq:rtrl} to obtain an unbiased approximation $\hat{G}_{t+1}$ at time $t+1$
\begin{align}
    \hat{G}_{t+1} = \frac{\partial \Fstate}{\partial \theta}(x_{t+1}, s_t, \theta) +
    \frac{\partial \Fstate}{\partial s}(x_{t+1}, s_t, \theta) \,\sbar{t} \otimes \thetabar{t}.
    \label{eq:ghat}
\end{align}
However, in general this is no longer rank-one.

To transform $\hat{G}_{t+1}$ into $\tilde{G}_{t+1}$, a rank-one unbiased
approximation, the following rank-one trick, introduced in \cite{DBLP:journals/corr/OllivierC15} is used:
\begin{proposition}
    \label{prop:rk1}
    Let $A$ be a real matrix that decomposes as 
    \begin{equation}
        A = \sum\limits_{i=1}^k v_i\otimes w_i.
    \end{equation}
    Let $\nu$ be a vector of $k$ independent random signs, and $\rho$ a
    vector of $k$ positive numbers. Consider the rank-one matrix
    \begin{equation}
    \tilde{A}\deq \left(\sum\limits_{i=1}^k
        \rho_i \nu_i
	v_i\right)\otimes\left(\sum\limits_{i=1}^k\frac{\nu_i
	    w_i}{\rho_i}\right)
    \end{equation}
    Then $\tilde A$ is an unbiased rank-one approximation of $A$: $\E_{\nu} \tilde{A}=A$.
\end{proposition}

The rank-one trick can be applied for any $\rho$. The choice of $\rho$
influences the variance of the approximation; choosing
\begin{equation}
    \rho_i = \sqrt{\norm{w_i}/\norm{v_i}}
\end{equation}
minimizes the variance of the approximation,
$\mathbb{E}\left[\|A-\tilde{A}\|_2^2\right]$
\cite{DBLP:journals/corr/OllivierC15}.

The UORO update is obtained by applying the rank-one trick twice to
\eqref{eq:ghat}.
First,
$\frac{\partial \Fstate}{\partial \theta}(x_{t+1}, s_t, \theta)$ is
reduced to
a rank one matrix, without variance minimization.
\footnote{
    Variance minimization is not used at this step, since computing $\sqrt{\frac{\|w_i\|}{\|v_i\|}}$
    for every $i$ is not scalable.
} Namely,
let $\nu$ be a vector of independant random signs; then, 
\begin{equation}
    \frac{\partial \Fstate}{\partial \theta}(x_{t+1}, s_t, \theta)= \mathbb{E}_{\nu}\left[\nu \otimes \nu\trans\,\frac{\partial \Fstate}{\partial \theta}(x_{t+1}, s_t, \theta)\right].
    \label{eq:rk1ftheta}
\end{equation}
This results in a rank-two, unbiased estimate of $\partial s_{t+1}/\partial \theta$
by substituting \eqref{eq:rk1ftheta} into \eqref{eq:ghat}
\begin{align}
    \frac{\partial \Fstate}{\partial s}&(x_{t+1}, s_t, \theta)\,\sbar{t}\otimes\thetabar{t} 
    + \nu \otimes \!\left(\nu\trans \, \frac{\partial \Fstate}{\partial
    \theta}(x_{t+1}, s_t, \theta)\right).
\end{align}
Applying Proposition~\ref{prop:rk1}
a second time to this rank-two estimate, with variance minimization, yields UORO's estimate $\tilde{G}_{t+1}$
\begin{align}
    \tilde{G}_{t+1} =
    \left( \rho_0\frac{\partial \Fstate}{\partial s}(x_{t+1}, s_t, \theta)\, \sbar{t} + \rho_1\,\nu\right)
    \otimes
    \left(\frac{\thetabar{t}}{\rho_0}+\frac{\nu}{\rho_1}\trans\,\frac{\partial \Fstate}{\partial \theta}(x_{t+1}, s_t, \theta)\right)
    \label{eq:gtilde}
\end{align}
which satisfies that $\E_\nu \,\tilde{G}_{t+1}$ is equal to
\eqref{eq:ghat}. (By elementary algebra, some
random signs that should appear in \eqref{eq:gtilde} cancel out.) Here
\begin{align}
    \rho_0 = \sqrt{\frac{\|\thetabar{t}\|}{\|\frac{\partial \Fstate}{\partial s}(x_{t+1}, s_t, \theta)\, \sbar{t}\|}},\quad
    \rho_1 = \sqrt{\frac{\|\nu\trans \, \frac{\partial \Fstate}{\partial \theta}(x_{t+1},s_t,\theta)\|}{\|\nu\|}}
\end{align}
minimizes variance of the second reduction.

The unbiased estimation \eqref{eq:gtilde} is rank-one and can be
rewritten as
$\tilde{G}_{t+1}=\sbar{t+1}\otimes \thetabar{t+1}$ with the update
\begin{align}
    \sbar{t+1} &\leftarrow \rho_0 \,\frac{\partial \Fstate}{\partial s}(x_{t+1}, s_t, \theta)\, \sbar{t} + \rho_1 \, \nu \label{eq:sbar}\\
    \thetabar{t+1} &\leftarrow \frac{\thetabar{t}}{\rho_0} +
    \frac{\nu\trans}{\rho_1} \, \frac{\partial \Fstate}{\partial \theta}(x_{t+1}, s_t, \theta).
    \label{eq:thetabar}
\end{align}

Initially, $\partial s_0/\partial \theta = 0$, thus $\sbar{0} = 0$, $\thetabar{0} = 0$ yield an
unbiased estimate at time $0$. Using this initial estimate, as well as the update rules
\eqref{eq:sbar}--\eqref{eq:thetabar}, an estimate of $\partial s_t/\partial \theta$ is obtained at all
subsequent times, allowing for online estimation of $\partial
\ell_t/\partial \theta$. Thanks to the construction above, by induction
all these estimates are unbiased.
\footnote{
    In practice, since $\theta$ changes during learning, unbiasedness
    only holds exactly in the limit of small learning rates. This is not
    specific to UORO as it also affects RTRL.
}

We are left to demonstrate that these update rules are scalably implementable.

\subsection{Implementation}
\label{sec:impl}
Implementing UORO requires maintaining the rank-one approximation and
the corresponding gradient loss estimate.

UORO's estimate of the loss gradient $\partial\ell_{t+1}/\partial_\theta$ at time $t+1$ is expressed by plugging into
\eqref{eq:lossgrad} the rank-one approximation $\partial s_t/\partial
\theta \approx \sbar{t}\otimes\thetabar{t}$, which results in
\begin{align}
    \left(\frac{\partial \ell_{t+1}}{\partial o}(o_{t+1}, \target{o}_{t+1})\,\frac{\partial \Fout}{\partial s}(x_{t+1}, s_t, \theta)\cdot\sbar{t}\right)\,\thetabar{t}
    +\frac{\partial \ell_{t+1}}{\partial o}(o_{t+1}, \target{o}_{t+1})\,\frac{\partial \Fout}{\partial \theta}(x_{t+1}, s_t, \theta).
    \label{eq:glossest}
\end{align}
Backpropagating $\partial \ell_{t+1}/\partial o_{t+1}$ once
through $\Fout$ returns\\
$\left(\partial \ell_{t+1}/\partial o_{t+1}\cdot \partial \Fout/\partial
x_{t+1}, \, \partial \ell_{t+1}/\partial o_{t+1}\cdot\partial
\Fout/\partial s_t, \,
 \partial \ell_{t+1}/\partial o_{t+1}\cdot\partial \Fout/\partial
 \theta\right)$, thus providing all necessary terms to compute
\eqref{eq:glossest}.

Updating $\sbar{}$ and $\thetabar{}$ requires applying \eqref{eq:sbar}--\eqref{eq:thetabar}
at each step. Backpropagating the vector of random signs $\nu$ once
through $\Fstate$ returns
$\left(\_,\_ , \nu\trans\,{\partial\Fstate(x_{t+1}, s_t, \theta)}/{\partial
\theta}\right)$, providing for \eqref{eq:thetabar}.

Updating $\sbar{}$ via \eqref{eq:sbar} requires computing $(\partial \Fstate/\partial s_t)\cdot \sbar{t}$.
This is computable numerically through 
\begin{equation}
    \frac{\partial \Fstate}{\partial s}(x_{t+1}, s_t, \theta) \cdot
    \sbar{t}=
    \lim\limits_{\varepsilon\to 0}\frac{\Fstate(x_{t+1}, s_t +
    \varepsilon\,\sbar{t}, \theta) - \Fstate(x_{t+1}, s_t,
    \theta)}{\varepsilon}
\end{equation}
computable through two applications of $\Fstate$. This operation is referred to as
tangent forward propagation \cite{tangprop} and can also often be computed
algebraically.

This allows for complete implementation of one step of UORO
(Alg.~\ref{alg:uoro}).  The cost of UORO (including running the model
itself) is three
applications of $\Fstate$, one application of $\Fout$, one
backpropagation through $\Fout$ and $\Fstate$, and a few elementwise
operations on vectors
and scalar products.

The resulting algorithm is detailed in Alg.~\ref{alg:uoro}.
$F.\textbf{forward}(v)$ denotes pointwise application of $F$
at point $v$, $F.\textbf{backprop}(v,\delta o)$ backpropagation of row vector
$\delta o$ through $F$ at point $v$, and $F.\textbf{forwarddiff}(v,\delta v)$
tangent forward propagation of column vector $\delta v$ through $F$
at point $v$. Notably, $F.\textbf{backprop}(v, \delta o)$ has the same
dimension as $v\trans$, e.g. $\Fout.\textbf{backprop}((x_{t+1}, s_t, \theta),
\delta o_{t+1})$ has three components, of the same dimensions as
$x_{t+1}\trans$, $s_t\trans$ and $\theta\trans$.

The proposed update rule for stochastic gradient descent \eqref{eq:sgd}
can be directly adapted to other optimizers, e.g. \emph{Adaptative
Momentum} (Adam) \cite{DBLP:journals/corr/KingmaB14} or \emph{Adaptative
Gradient}~\cite{Duchi:EECS-2010-24}. Vanilla stochastic gradient descent
(SGD) and Adam are used hereafter.  In Alg.~\ref{alg:uoro}, such optimizers are denoted by
$\SGDOpt$ and the corresponding parameter update given current parameter
$\theta$, gradient estimate $g_t$ and learning rate $\eta_t$ is denoted
$\SGDOpt.\textbf{update}(g_t, \eta_t, \theta)$.

\begin{algorithm}

\caption{--- One step of UORO (from time $t$ to $t+1$)}
\label{alg:uoro}

\begin{algorithmic}

\STATE {\bfseries Inputs:}\\ 

\begin{itemize}[label=--,noitemsep]

\item $x_{t+1}$, $\target{o}_{t+1}$, $s_t$ and $\theta$: input, target,
previous recurrent state, and parameters
\item $\sbar{t}$ column vector of size $\mathit{state}$,
$\thetabar{t}$ row vector of size $\mathit{params}$ such that
$\E\, \sbar{t}\otimes \thetabar{t} = \partial
s_t/\partial \theta$
\item $\SGDOpt$ and $\eta_{t+1}$: stochastic optimizer and its learning rate

\end{itemize}

\STATE {\bfseries Outputs:}\\

\begin{itemize}[label=--,noitemsep]

\item $\ell_{t+1}$, $s_{t+1}$ and $\theta$: loss, new recurrent state, and updated parameters
\item $\sbar{t+1}$ and $\thetabar{t+1}$ such that $\E\, \sbar{t+1}\otimes \thetabar{t+1} = \partial s_{t+1}/\partial \theta$
\item $\gbar{t+1}$ such that $\E \,\gbar{t+1}=\partial \ell_{t+1}/\partial \theta$

\end{itemize}

    \STATE \comment{compute next state and loss}
\STATE
$
s_{t+1} \gets \Fstate.\textbf{forward}(x_{t+1}, s_t,\theta),\quad
o_{t+1} \gets \Fout.\textbf{forward}(x_{t+1}, s_t, \theta)$
\STATE $\ell_{t+1} \gets \ell(o_{t+1}, \target{o}_{t+1})$\;
\STATE
\STATE \comment{compute gradient estimate}
\STATE $(\_, \delta s, \delta \theta) \leftarrow
\Fout.\textbf{backprop}\left((x_{t+1}, s_t, \theta), {\displaystyle \frac{\partial
\ell_{t+1}}{\partial o_{t+1}}}\right)$\;
\STATE $\gbar{t+1} \leftarrow (\delta s \cdot \tilde{s}_t) \,\tilde{\theta}_t + \delta \theta$\;
\STATE
\STATE \comment{prepare for reduction}
\STATE Draw $\nu$, column vector of random signs $\pm1$ of size $\mathit{state}$\\
\STATE $\sbar{t+1}\leftarrow \Fstate.\textbf{forwarddiff}((x_{t+1}, s_t,
	\theta),(0, \sbar{t}, 0))$\; 
\STATE $(\_, \_, \delta \theta_g) \leftarrow \Fstate.\textbf{backprop}((x_{t+1},
	s_t, \theta), \nu\trans)$\;
\STATE
\STATE \comment{compute normalizers}
\STATE $\rho_0 \leftarrow \displaystyle
	\sqrt{\frac{\|\thetabar{t}\|}{\norm{\sbar{t+1}}+\varepsilon}}+\epsilon\,, \quad 
        \rho_1 \leftarrow \sqrt{\frac{\|\delta \theta_g\|}{\|\nu\|+\varepsilon}}+\epsilon$ with $\varepsilon=10^{-7}$\;
\STATE
\STATE \comment{reduce}
\STATE $\sbar{t+1} \leftarrow \rho_0\, \sbar{t+1} + \rho_1\, \nu , \quad
\displaystyle\thetabar{t+1} \leftarrow \frac{\thetabar{t}}{\rho_0} + \frac{\delta \theta_g}{\rho_1}$\;
\STATE \comment{update $\theta$}
\STATE $\SGDOpt.\textbf{update}(\gbar{t+1}, \eta_{t+1}, \theta)$
\end{algorithmic}
\end{algorithm}

\subsection{Memory-$T$ UORO and rank-$k$ UORO}
\label{sec:extension}
The unbiased gradient estimates of UORO come at the price of noise
injection via $\nu$.
This requires
smaller learning rates. To reduce noise, UORO can be also used on top of
truncated BPTT so that recent gradients are computed exactly.

Formally, this just requires applying Algorithm~\ref{alg:uoro} to a new
transition function $F^T$ which is just $T$ consecutive steps of the
original model $F$. Then the backpropagation operation in
Algorithm~\ref{alg:uoro} becomes a backpropagation over the last $T$
steps, as in truncated BPTT. The loss of one step of $F^T$ is the sum of the losses of the last
$T$ steps of $F$, namely $\ell_{t+1}^{t+T} \deq \sum\limits_{k=t+1}^{t+T}
\ell_k$. Likewise, the forward tangent propagation is performed through
$F^T$. This way, we obtain an unbiased gradient estimate in which the
gradients from the last $T$ steps are computed exactly and incur no
noise.

The resulting algorithm is referred to as memory-$T$ UORO. Its scaling in $T$ is
similar to $T$-truncated BPTT, both in terms of memory and computation.  In the
experiments below, memory-$T$ UORO reduced variance early on,
but did not significantly impact later performance.

The noise in UORO can also be reduced by using higher-rank gradient
estimates (rank-$r$ instead of rank-$1$), which amounts to maintaining
$r$
distinct values of $\sbar{}$ and $\thetabar{}$ in
Algorithm~\ref{alg:uoro} and averaging the resulting values of $\gbar{}$.
We did not exploit this possibility in the experiments below, although
$r=2$ visibly reduced variance in preliminary tests.

\section{Experiments illustrating truncation bias}
\label{sec:exp}

The set of experiments below aims at displaying
specific cases where the biases from truncated BPTT are
likely to prevent convergence of learning. On this test
set, UORO's unbiasedness provides steady convergence,
highlighting the importance of unbiased estimates for
general recurrent learning.

\paragraph{Influence balancing.}
The first test case exemplifies learning of a scalar parameter $\theta$ which has
a positive influence in the short term, but a negative one in the long run. Short-sightedness
of truncated algorithms results in abrupt failure, with the parameter
exploding in the wrong direction, even with truncation lengths exceeding the
temporal
dependency range by a factor of $10$ or so.

\begin{figure}[h!]
    \vspace{1.5em}
    \begin{subfigure}[b]{.42\linewidth}
        \centering
        \vspace{-0.75cm}
        \includegraphics[scale=.6]{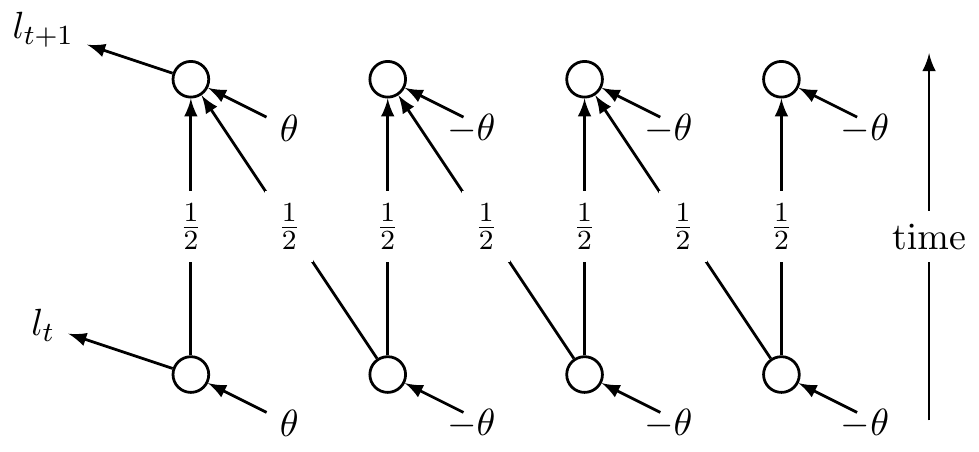}
        \caption{Influence balancing, $4$ units, $3$ minus.}
        \label{fig:infbal}
    \end{subfigure}
    \begin{subfigure}[b]{.35\linewidth}
        \centering
        \begin{BVerbatim}
[a]eecbe[a]
[j]fbfjd[j]
[c]bgddc[c]
[d]gjhai[d]
[e]iaghb[e]
[h]bigaj[h]
        \end{BVerbatim}
        \caption{Distant brackets $(1,5,10)$.}
        \label{fig:distantsample}
    \end{subfigure}
    \begin{subfigure}[b]{.2\linewidth}
        \centering
        \begin{BVerbatim}
aaaaaa
bbbbbb
aaaaaaaaaaaaaaaa
bbbbbbbbbbbbbbbb
aaaaaaaa
bbbbbbbb
        \end{BVerbatim}
        \caption{$a^nb^n(1,32)$.}
        \label{fig:anbnsample}
    \end{subfigure}
    \caption{Datasets.}
    \label{fig:synthtext}
    \vspace{-1em}
\end{figure}

Consider the linear dynamics
\begin{equation}
    s_{t+1} = A \, s_t + (\theta, \ldots, \theta, -\theta, \ldots, -\theta)\trans
\end{equation}
with $A$ a square matrix of size $n$ with $A_{i,i} = 1/2$,
$A_{i,i+1}=1/2$, and $0$ elsewhere; $\theta \in \real$ is a scalar
parameter. 
The second term has $p$ positive-$\theta$ entries and $n-p$ negative-$\theta$ entries.
Intuitively, the effect of $\theta$ on a unit diffuses to shallower units
over time (Fig.~\ref{fig:infbal}).
Unit $i$ only feels the effect of $\theta$ from unit $i+n$
after $n$ time steps, so the intrinsic time scale of the system is
$\approx n$. The loss considered is a target on the shallowest
unit $s^1$,
\begin{equation}
    \ell_t = {\textstyle \frac12} (s^1_t - 1)^2.
\end{equation}

Learning is performed online with vanilla SGD, using
gradient estimates either from UORO or $T$-truncated BPTT with various $T$.
Learning rates are of the form
$\eta_t=\frac{\eta}{1 + \sqrt{t}}$ for suitable values of $\eta$.

\begin{figure*}[ht!]
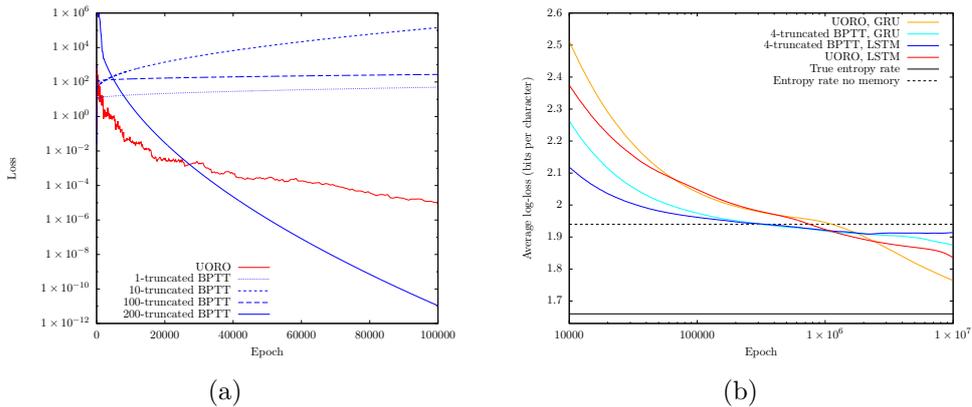

\vspace{-1.2em}
    \begin{subfigure}{.4\textwidth}
        \scalebox{.4}{\input{bptt-f-arxiv.tex}}
        \caption{}
        \label{fig:bpttf}
    \end{subfigure}\hspace{2em}
    \begin{subfigure}{.4\textwidth}
        \scalebox{.4}{\input{distant-bracket-arxiv.tex}}
        \caption{}
        \label{fig:distantbrackets}
    \end{subfigure}
    \caption{(a)Results for influence balancing with $23$ units and $13$
    minus; note the vertical log scale.
    (b)Learning curves on distant brackets $(1, 5, 5, 10)$.}
\end{figure*}

As shown in Fig.~\ref{fig:bpttf}, UORO solves the problem while
$T$-truncated BPTT
fails to converge for any learning rate, even for truncations $T$
largely above $n$. Failure is caused by ill balancing of time
dependencies: the influence of $\theta$ on the loss is estimated with the
wrong sign due to truncation.
For $n=23$ units, with $13$ minus signs, truncated BPTT requires a truncation
$T\geq 200$ to converge.

\paragraph{Next-character prediction.}
The next experiment is character-level synthetic text prediction:
the goal is to train a recurrent model to predict the $t+1$-th character of a text given the 
first $t$ online, with a single pass on the data sequence. 

A single layer of $64$ units, either GRU or LSTM, is used to output a
probability vector for the next character. The cross entropy criterion is
used to compute the loss.


At each time $t$ we plot the cumulated loss per character on the first
$t$ characters, $\frac1t\sum_{s=1}^t\ell_s$. (Losses for individual
characters are quite noisy, as not all characters in the sequence are equally difficult to
predict.) This would be the
compression rate in bits per character if the models were used as
online compression algorithms on the first $t$ characters. In addition,
in Table~\ref{fig:anbnsum} we report a ``recent'' loss on the last
$100,000$ characters, which is more representative of the model at the
end of learning.

Optimization was performed using Adam with the default setting $\beta_1 = 0.9$ and $\beta_2 =
0.999$, and a decreasing learning rate $\eta_t = \frac{\gamma}{1+\alpha
\sqrt{t}}$, with $t$ the number of characters processed. As convergence
of UORO requires
smaller learning rates than truncated BPTT, this favors UORO. Indeed
UORO can fail to converge with non-decreasing learning rates, due to
its stochastic nature.

\begin{figure*}[h!]
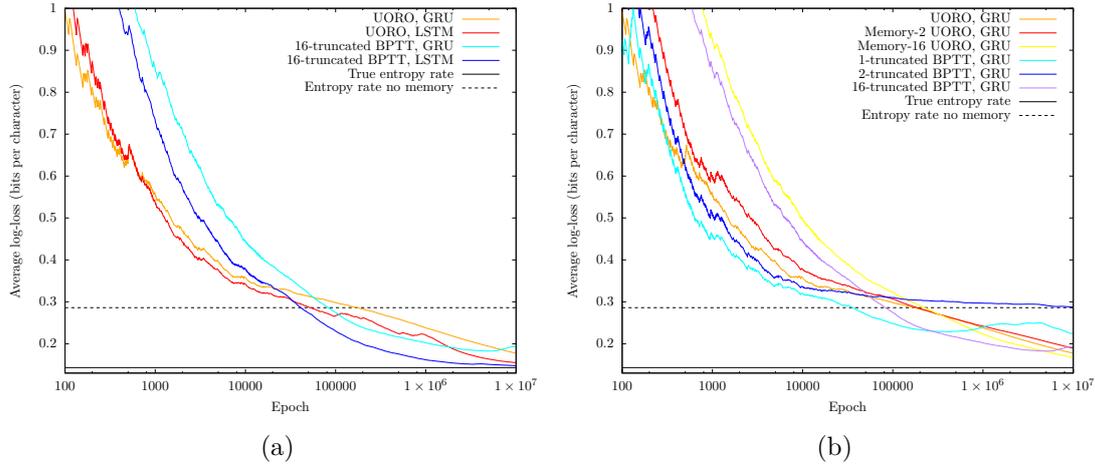

    \begin{subfigure}{.49\textwidth}
        \scalebox{.47}{\input{anbn1-arxiv.tex}}
        \caption{}
    \end{subfigure}
    \begin{subfigure}{.49\textwidth}
        \scalebox{.47}{\input{anbn2-arxiv.tex}}
        \caption{}
        \label{fig:anbnnoise}
    \end{subfigure}
\caption{Learning curves on $a^nb^n_{(1,32)}$}
\label{fig:anbnplot}
\end{figure*}

\subparagraph{Distant brackets dataset $(s, k, a)$.}
The distant brackets dataset is generated by repeatedly outputting a left bracket,
generating $s$ random characters from an alphabet of size $a$, outputting a right
bracket,  generating $k$ random
characters from the same alphabet, repeating the same first $s$ characters between
brackets and finally outputting a line break. A sample is shown in Fig.~\ref{fig:distantsample}. 

UORO is compared to $4$-truncated BPTT.
Truncation is
deliberately shorter than the inherent time range of the data, to illustrate
how bias can penalize learning if the inherent time range is unknown a
priori.  
The results are given in Fig.~\ref{fig:distantbrackets} (with learning
rates using 
$\alpha = 0.015$ and
$\gamma = 10^{-3}$).
UORO beats
$4$-truncated BPTT in the
long run, and succeeds in reaching near optimal behaviour both with GRUs
and LSTMs. Truncated BPTT remains stuck near a memoryless optimum with
LSTMs; with GRUs it keeps learning, but at a slow rate.
Still, 
 truncated BPTT displays faster early
convergence. 

\subparagraph{$a^nb^n(k, l)$ dataset}
The $a^nb^n(k, l)$ dataset tests memory and counting \cite{gers2001long};
it is generated by repeatedly picking a random
number $n$ between $k$ and $l$, outputting a string of $n$ $a$'s, a line break,
$n$ $b$'s, and a line break (see Fig.~\ref{fig:anbnsample}).  The
difficulty lies in matching the number of
$a$'s and $b$'s.

\begin{table}[h!]
    \centering
    \vspace{-1.2em}
    \caption{Averaged loss on the $10^5$ last iterations on $a^nb^n(1,32)$.}
\begin{tabular}{l l l l}
    & Truncation & LSTM & GRU \\
    \hline
    \multirow{3}{*}{UORO}
    & No memory (default)& $0.147$ & $0.155$\\
    & Memory-2 & $0.149$ & $0.174$\\
    & Memory-16 & $0.154$ & $0.149$\\
    \hline
    \multirow{3}{*}{Truncated BPTT}
    & 1 & $0.178$ & $0.231$ \\
    & 2 & $0.149$ & $0.285$ \\
    & 16 & $0.144$ & $0.207$\\
    \hline
\end{tabular}
\label{fig:anbnsum}
\end{table}

Plots for a few setups are given in Fig.~\ref{fig:anbnplot}. 
The learning rates used $\alpha = 0.03$ and $\gamma = 10^{-3}$.

Numerical results at the end of training are given in
Table~\ref{fig:anbnsum}. For reference, the true entropy rate is $0.14$
bits per character, while the entropy rate of a model that does not
understand that the numbers of $a$'s and $b$'s coincide would be double, $0.28$ bpc.

Here, in every setup, UORO reliably converges and reaches near optimal performance.
Increasing UORO's range does not significantly improve results: providing
an unbiased estimate is enough to provide reliable convergence in this case.
Meanwhile, truncated BPTT performs inconsistently.
Notably, with GRUs, it
either converges to a poor local optimum corresponding to no understanding
of the temporal structure, or exhibits gradient reascent in the long run.
Remarkably, with LSTMs rather than GRUs, $16$-truncated BPTT reliably reaches optimal
behavior on this problem even
with biased gradient estimates.

%


\section*{Conclusion}
\label{sec:concl}
We introduced UORO, an algorithm for training recurrent neural
networks in a streaming, memoryless fashion. UORO is easy to implement,
and requires as little computation time as truncated BPTT, at the cost of noise injection.
Importantly, contrary to most other approaches, UORO scalably provides unbiasedness 
of gradient estimates. Unbiasedness is of paramount importance in the
current theory of stochastic gradient descent.

Furthermore, UORO is experimentally shown to benefit from its
unbiasedness, converging even in cases where truncated BPTT fails to reliably achieve
good results or diverges pathologically.

\cleardoublepage
\bibliography{uoro}
\bibliographystyle{alpha}
\end{document}